\documentclass[runningheads]{llncs}
\usepackage[T1]{fontenc}
\usepackage{graphicx}
\usepackage{acro}
\usepackage{xcolor}
\usepackage{subcaption}
\usepackage[switch]{lineno}
\usepackage{multirow}
\usepackage{multicol}
\usepackage{booktabs}
\usepackage{amssymb}
\usepackage{bm}
\usepackage{algorithm}
\usepackage{algorithmicx}
\usepackage{algpseudocode}
\usepackage{soul}
\usepackage{amsmath} 
\usepackage{dsfont}
\usepackage{hyperref}

\DeclareAcronym{DOF}{
short=DoF,
long=degrees of freedom}

\DeclareAcronym{DH}{
short=DH,
long=Denavit–Hartenberg}

\DeclareAcronym{BFGS}{
short= BFGS,
long= Broyden–Fletcher–Goldfarb–Shanno}

\DeclareAcronym{L-BFGS}{
short= L-BFGS,
long= Limited-memory Broyden–Fletcher–Goldfarb–Shanno}

\DeclareAcronym{L-BFGS-B}{
short= L-BFGS-B,
long= Limited-memory Broyden–Fletcher– Goldfarb–Shanno-Bounded}

\DeclareAcronym{DFP}{
short=DFP,
long=Davidon–Fletcher –Powell}

\DeclareAcronym{FK}{
short=FK,
long=forward kinematics}

\DeclareAcronym{IK}{
short=IK,
long=inverse kinematics}

\DeclareAcronym{GPU}{
short=GPU,
long=Graphics Processing Unit}

\DeclareAcronym{PDO-IK}{
short=PDO-IK,
long=propagative distance optimization for constrained inverse kinematics}

\DeclareAcronym{CoM}{
short=CoM,
long=center of mass}

\DeclareAcronym{R-TR}{
short=R-TR,
long=Riemannian Trust Region}

\DeclareAcronym{R-CG}{
short=R-CG,
long=Riemannian Conjugate Gradient}

\begin{document}
\title{Propagative Distance Optimization for Constrained Inverse Kinematics}
\author{
Yu Chen\inst{1} \and
Yilin Cai\inst{1} \and
Jinyun Xu\inst{1} \and \\
Zhongqiang Ren\inst{2} \and
Guanya Shi\inst{1} \and
Howie Choset\inst{1}
}
\authorrunning{Yu et al.}
\institute{
Carnegie Mellon University, Pittsburgh, PA 15213 USA \and
Shanghai Jiao Tong University, Shanghai, 200240 China
}
\maketitle              % typeset the header of the contribution
\begin{abstract}{
% This paper investigates a constrained inverse kinematic (IK) problem that seeks a feasible configuration of an articulated robot under various constraints such as joint limits and obstacle collision avoidance.
% This problem is often solved numerically via iterative local optimization, which however suffers from non-linearity caused by the joint angle parametrization.
% Distance-based methods take an alternative view by formulating IK as an optimization over the distances among points attached to the robot and the obstacles, and have demonstrated unique advantages.
% Currently, distance-based methods still suffer from low computational efficiency and often fail to find a feasible solution especially in cluttered environments.
% This paper proposes a new method called \ac{PDO-IK}, which captures and leverages the kinematics of the robot in the distance-based formulation and expedites the optimization.
% Test results show \ac{PDO-IK} runs up to two orders of magnitude faster than the existing distance-based methods, and achieves up to three times higher success rates than the conventional joint angle optimization.
% The high runtime efficiency of \ac{PDO-IK} allows the real-time computation (10$-$1500 Hz) and enables a simulated humanoid with 19 DOFs to avoid moving obstacles, which is otherwise hard to achieve with the baselines.
% This paper investigates a local optimization problem for constrained inverse kinematics (IK) that seeks a feasible configuration of an articulated robot under various constraints.
This paper investigates a constrained inverse kinematic (IK) problem that seeks a feasible configuration of an articulated robot under various constraints such as joint limits and obstacle collision avoidance. 
Due to the high-dimensionality and complex constraints, this problem is often solved numerically via iterative local optimization.
Classic local optimization methods take joint angles as the decision variable, which suffers from non-linearity caused by the trigonometric constraints.
Recently, distance-based IK methods have been developed as an alternative approach that formulates IK as an optimization over the distances among points attached to the robot and the obstacles.
Although distance-based methods have demonstrated unique advantages, they still suffer from low computational efficiency, since these approaches usually ignore the chain structure in the kinematics of serial robots.
This paper proposes a new method called \ac{PDO-IK},  which captures and leverages the chain structure in the distance-based formulation and expedites the optimization by computing forward kinematics and the Jacobian propagatively along the kinematic chain.
Test results show that \ac{PDO-IK} runs up to two orders of magnitude faster than the existing distance-based methods under joint limits constraints and obstacle avoidance constraints.
It also achieves up to three times higher success rates than the conventional joint-angle-based optimization methods for IK problems.
The high runtime efficiency of \ac{PDO-IK} allows the real-time computation (10$-$1500 Hz) and enables a simulated humanoid robot with 19 degrees of freedom (DoFs) to avoid moving obstacles, which is otherwise hard to achieve with the baselines.
}

% \guanya{\begin{itemize}
%     \item This paper studies XXX (define the problem and emphasize disability)
%     \item Due to the complex constraints, this problem is often solved by iterative local optimization.
%     \item In iterative local optimization, classic approaches (e.g., XXX [], XXX []) take joint angles as the decision variable, which suffers from XXX (explain using few words).
%     \item Recently, distance-based methods (xxx []) have been developed by using XXX (few words) as XXX, which significantly XXX (advantages compared to classic approaches).
%     \item Although being a promising alternative, distance-based suffers from efficiency because YYYYYY (short sentence for YYY).
%     \item In this work, we propose ZZZ, a distance-based XXX that (your idea to kill YYYYY). (one sentence for insight or explaining why it is much faster)
%     \item Performance. Broader impact.
% \end{itemize}}

\end{abstract}
\section{Introduction}\label{sec:intro}
For an articulated robot consisting of rigid links and revolute joints, the \ac{IK} problem seeks joint angles (i.e., a configuration) such that the end effector(s) reach a given pose, which is a fundamental problem in robotics.
This paper focuses on the constrained \ac{IK} that requires finding a feasible configuration under various constraints, including joint angle limits, and collision avoidance of workspace obstacles.
Constrained \ac{IK} can only be solved analytically for some specific robots.
For the general case, constrained \ac{IK} is usually formulated as constrained optimization problems and solved numerically via iterative local optimization.
% However, solving constrained \ac{IK} with a local optimizer is challenging. 
This local optimization often suffers from non-linearity due to the kinematic model of the robot, i.e., the mapping from the joint space, commonly represented with {\it angles}, to the task space, typically represented in the {\it Euclidean space} \cite{maric2021riemannian}.
In particular, the kinematic model leads to complicated trigonometric constraints, which can trap the optimization in a local optimum that is highly sub-optimal or even infeasible, especially in the presence of high \ac{DOF} and cluttered workspace with many obstacles.

To address this challenge, an important class of methods in the literature is to eliminate the trigonometric mapping in the kinematic model
% , prior works have proposed various kinematic algorithms based on 
by using distance-based optimization \cite{porta2005inverse,han2006inverse,maric2020inverse,maric2021riemannian,giamou2022convex,maric2020inverse2,porta2005branch}.
Instead of optimizing the joint angles, distance-based methods attach points to the robot and the obstacles, and reformulate the constrained \ac{IK} as an optimization over the distances among these points.
% represent the robot and environment with the invariant distances between points attached to the robot, target frames, and obstacles.
To name a few, 
Josep et al. \cite{porta2005inverse} formulated the kinematics of 6-\ac{DOF} serial robots using a distance matrix \cite{crippen1988distance} and solve the \ac{IK} with matrix completion leveraging Cayley-Menger determinant \cite{sippl1986cayley}. 
% Han et al. \cite{han2006inverse} parameterized the \ac{IK} problem of serial robots using a combination of anchored diagonal lengths and triangle orientations.
% Although their method can be applied to serial robots with arbitrary \ac{DOF}s, they ignore joint limits and possible collisions.
Marić and Giamou et al captures the sparsity of the distance matrix \cite{maric2020inverse2} and use sparse bounded-degree sum of squares relaxations \cite{weisser2018sparse} to solve \ac{IK} for spherical joint robots without considering collisions.
Marić et al. proposed a distance-geometric framework called \ac{R-TR} \cite{maric2021riemannian} to solve the constrained \ac{IK} by optimizing the distance matrix using Riemannian optimization.
% Their approach demonstrates more effective performance in constrained \ac{IK} computations compared to angle-based algorithms, particularly in managing joint limits and collision avoidance.

Despite these advancements, distance-based methods still suffer from low computational efficiency.
% and often fail to find a feasible solution especially when the environment is cluttered with many obstacles.
% in complex, unstructured environments. Some works, such as \cite{porta2005inverse} and \cite{han2006inverse}, lack the ability to handle important constraints in \ac{IK} problems. While other solvers, such as \ac{R-TR}, perform slow speed and are likely to fail in finding feasible kinematic solutions when the number of obstacles increases.
This paper proposes a new distance-based \ac{IK} method called \ac{PDO-IK}.
Our key insight is that: Most existing distance-based methods for \ac{IK} ignore the chain structure of the kinematics of serial robots after the reformulation, and purely focus on optimizing the distances among the points.
In contrast, \ac{PDO-IK} derives a new kinematic model based on the distance between points attached to the robot that captures and leverages such chain structures. \ac{PDO-IK} computes the forward kinematics and Jacobian propagatively along the kinematic chain, which decomposes the kinematics model into a set of serially connected \textit{units} and iteratively solve for one unit after another along the chain.
In our formulation, such \textit{units} are described using the points attached on each frame of robot link.
This introduces two advantages: First, the computation of any unknown variable can re-use the variables that have already been computed in its neighbouring unit. Second, the matrix of the distances between points is sparse, which reduces the amount of variables to compute. These advantages allow fast computation of the forward kinematics and Jacobians, and thus expedite the overall optimization.
% Intuitively, \ac{PDO-IK} respects the fact that points attached to a link of the robot is only affected by the points on the previous and the next links 
In particular, our technical contributions include both (i) a novel distance-based formulation of the constrained \ac{IK}, and (ii) an optimization algorithm using augmented Lagrangian based on the proposed formulation and the analysis of its runtime complexity.

% We propose \ac{PDO-IK}, a kinematics algorithm that optimize over distances. As shown in Fig.~\ref{fig:kinematics-model}, \ac{PDO-IK} represents the robot with points attached to the robot joints, models the environment obstacles with clustered point clouds, and encodes these points into a unified tree structure. The vertices of the tree are the positions of the points representing the robot or obstacles, and the edges are weighted by the distances between vertices. Joint angle limits, end effector pose, and collision avoidance are all converted to edge distance constraints in \ac{PDO-IK}. To efficiently solve the constrained \ac{IK} problem, \ac{PDO-IK} models the kinematic chain by constructing \ac{DH} matrices, which enable computing forward kinematics and Jacobians in a dynamic programming manner. Based on these parameterizations and structure formulations, \ac{PDO-IK} solves \ac{IK} problems with a Newton-based local optimizer. We theoretically prove that our algorithm has linear complexity with respect to the number of vertices in the tree. Numerical simulation results across various robot arms demonstrate that our algorithm outperforms baseline distance-based \ac{IK} algorithms in efficiency and effectiveness, especially in complex environments. Simulation experiments on a humanoid robot show that our algorithm can be generalized to diverse kinds of constraints, such as \ac{CoM} limitations.

For verification, we compare our \ac{PDO-IK} against both a joint angle-based optimization method and some recent distance-based methods~\cite{maric2021riemannian} as baselines in various settings.
The results show \ac{PDO-IK} can often double or triple the success rates (i.e., finding a feasible solution within a runtime limit) of the joint angle method, and run up to two orders of magnitude faster than the existing distance-based methods, especially in cluttered workspace.
In addition, \ac{PDO-IK} demonstrates better numerical robustness than the baselines in the sense that \ac{PDO-IK} can achieve a small numerical error tolerance that is below $10^{-4}$, while the error tolerance of the baselines is often larger than $10^{-3}$.
Finally, we show the generalization capability of \ac{PDO-IK} by applying it on a humanoid robot (19 \ac{DOF}s) with the additional position constraints on the \ac{CoM}. The runtime efficiency of \ac{PDO-IK} allows the real-time computation (10$-$1500 Hz) and enables the robot to avoid dynamic obstacles, which is otherwise hard to achieve with the baselines.

\section{Preliminaries}\label{sec:prel}
Robot kinematics describes the relationship between the configuration space $\mathcal{C}$ and the task space $\mathcal{T}$. The mapping $F : \mathcal{C} \rightarrow \mathcal{T}$ is the \ac{FK}, while its inverse $F^{-1} : \mathcal{T} \rightarrow \mathcal{C}$ is the \ac{IK}. In this paper, we focus on the end-effector pose objective defined in $\mathcal{T}$, joint limits as box constraints defined in $\mathcal{C}$, and collision avoidance constraints.

We consider a serial robot with $M$ revolute joints \footnote{Tree-structure robots and parallel robots can also be handled by adding extra constraints. An example is tested in Sec.~\ref{subsec:exp3}.}. We use $i$ as the index for joints and links. For the robot bodies, $i = 1, \dots, M$ and we define link 0 as the fixed base. Joint $i$ lies between link $i-1$ and link $i$. We use vector $\boldsymbol{\theta} = [\theta_1, \dots, \theta_M] \in \mathbb{R}^M$ to represent the joint angles with $\theta_i$ being the revolute angle of joint $i$. We also define unit vectors $\mathbf{x}_i$, $\mathbf{y}_i$, and $\mathbf{z}_i$ to denote the $x$-, $y$-, and $z$-axes of frame $ \mathcal{F}_i $, which is the coordinate frame attached to link $i$. The obstacles in the environment are considered as clusters of points, with a total of $N$ points. $\lceil \cdot \rceil$ denotes the ceiling function. $\sum (\cdot)$ denotes the sum of all elements within a vector or matrix.

\vspace{-3mm}
\subsection{Denavit–Hartenberg Parameters}\label{subsec:prel-DH}
\begin{figure}[t]
    \centering
    \includegraphics[width=1\linewidth]{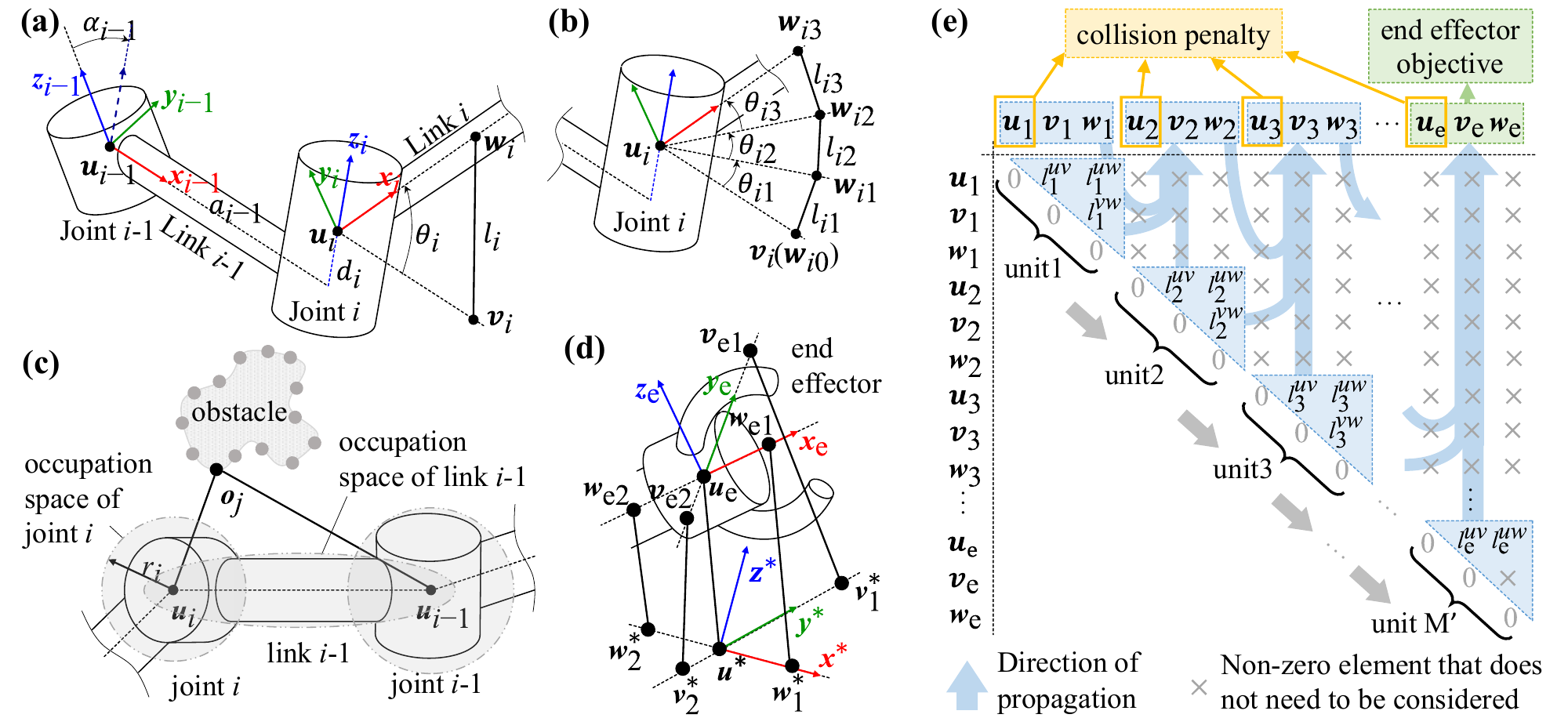}
    \caption{Kinematics model, constraints, and objective under distance-based representation, as well as the propagation structure in our method. (a) The kinematic chain of a linkage of revolute joints. (b) Joint angle decomposition. (c) Collision avoidance constraint. (d) End effector pose objective. (e) The propagation structure in a diagonal matrix of distances in the forward rollout. The propagation in Jacobian computation follows the inverse direction of the forward rollout.}
    \label{fig:kinematics-model}
\vspace{-4mm}
\end{figure}

The kinematic chain formulation in this paper builds upon the proximal \ac{DH} convention \cite{craig2006introduction}. As shown in Fig.~\ref{fig:kinematics-model}a, we attach the origin $\mathbf{o}_i$ of frame $\mathcal{F}_i$ to the revolute axis of joint $i$. The direction of the axis $\mathbf{z}_i$ aligns with the revolute axis of joint $i$. $\mathbf{x}_i$ is perpendicular to and intersects $\mathbf{z}_i$ and $\mathbf{z}_{i+1}$. $\mathbf{y}_i$ is defined by the right-hand rule with $\mathbf{x}_i$ and $\mathbf{z}_i$. The transformation from frame $\mathcal{F}_{i-1}$ to frame $\mathcal{F}_i$ is:
\vspace{-3mm}
\begin{equation}
   ^{i-1}\mathbf{T}_i = f(\theta_i; \alpha_{i-1}, a_{i-1}, d_i) = \left[
                        \begin{array}{cccc}
                            c_\theta \quad & -s_\theta \quad & 0 \quad & a_{i-1} \\
                            s_\theta c_\alpha \quad & c_\theta c_\alpha \quad & -s_\alpha \quad & d_i s_\alpha \\
                            s_\theta s_\alpha \quad & c_\theta s_\alpha \quad & c_\alpha \quad & d_i c_\alpha \\
                            0 \quad & 0 \quad & 0 \quad & 1
                        \end{array}
                        \right]
    \label{eq:DH_matrix}
\vspace{-2mm}
\end{equation}
where $i \geq 1$, $c_\theta = \cos{\theta_i}$, $s_\theta = \sin{\theta_i}$, $c_\alpha = \cos{\alpha_{i-1}}$, $s_\alpha = \sin{\alpha_{i-1}}$, $a_{i-1}$ is the revolute radius of $\mathcal{F}_i$ to axis $\mathbf{z}_{i-1}$, $\alpha_{i-1}$ is the angle from $\mathbf{z}_{i-1}$ to $\mathbf{z}_i$ about common normal, $d_i$ is the offset of $\mathcal{F}_i$ to $\mathcal{F}_{i-1}$ along $\mathbf{z}_i$, and $\theta_i$ is the revolute angle from $\mathbf{x}_{i-1}$ to $\mathbf{x}_i$ about $\mathbf{z}_i$.

\vspace{-3mm}
\subsection{Quasi-Newton Method}\label{subsec:quasi-Newton}
Quasi-Newton method \cite{dennis1977quasi} is a variant of Newton's method when the Hessian is unavailable or expensive to compute, which iteratively finds the minimum of a function $f(x)$ by finding the root of its Jacobian $\nabla f$. Quasi-Newton methods approximate Hessian in each iteration $k$ given $f(x_k)$ and $\nabla f(x_k)$. Popular quasi-Newton methods include \ac{L-BFGS} \cite{liu1989limited}, Broyden's method \cite{broyden1965class}, etc.

\section{Kinematics and Constraints Formulation}\label{sec:model}
\ac{PDO-IK} represents joint space, task space, and constraints with Euclidean distances between points attached on the robot and obstacles. This section primarily discusses our formulation of the robot kinematics chain (Sec.~\ref{subsec:kinematic-chain}), joint limit constraints (Sec.~\ref{subsec:joint-limit}), collision avoidance constraints (Sec.~\ref{subsec:coll_avoid}), and end effector pose objective (Sec.~\ref{subsec:ee-pose}).

\vspace{-3mm}
\subsection{Kinematic Chain}\label{subsec:kinematic-chain}
We formulate the kinematic chain by modifying the proximal \ac{DH} convention, replacing the joint angles with Euclidean distances between points. As shown in Fig.~\ref{fig:kinematics-model}a, we first attach points to the link frames $\mathcal{F}_i$ and then re-parameterize the proximal DH matrix with the Euclidean distances among points.

We attach three points on $\mathcal{F}_i$. The first point $\mathbf{u}_i$ is attached to the origin $\mathbf{o}_i$ to represent the spatial position of joint $i$. The second point $\mathbf{v}_i$ is of distance $l_i^{uv}$ away from $\mathbf{u}_i$ in the direction of $\mathbf{x}_{i-1}$. The third point $\mathbf{w}_i$ is of distance $l_i^{uw}$ away from $\mathbf{u}_i$ in the direction of $\mathbf{x}_i$. The distance between $\mathbf{v}_i$ and $\mathbf{w}_i$, denoted as $l_i^{vw}$, is determined by $\theta_i$ using the Law of Cosines. The positions of these points and their confinements describe the robot's structure and motion:
\vspace{-5mm}
\begin{subequations}
\begin{multicols}{2}\noindent
    \begin{equation}
        \mathbf{u}_i = \mathbf{o}_i
    \vspace{-2mm}
    \end{equation}
    \begin{equation}
        \mathbf{v}_i - \mathbf{u}_i = l_i^{uv}\mathbf{x}_{i-1}
    \vspace{-2mm}
    \end{equation}
\end{multicols}
\begin{multicols}{2}\noindent
    \begin{equation}
        \mathbf{w}_i - \mathbf{u}_i = l_i^{wv}\mathbf{x}_{i}
    \end{equation}
    \begin{equation}
        2 l_i^{uv} l_i^{uw} \cos{\theta_i} = {l_i^{uv}}^2 + {l_i^{uw}}^2 - {l_i^{vw}}^2
        \label{eq:cos_law}
    \end{equation}
\end{multicols}
\end{subequations}
We define the squared distance $L_i={l_i^{vw}}^2$ so that $\cos{\theta_i} \propto L_i$. For simplicity, we let $l_i^{uv} = l_i^{uw} = 1/\sqrt{2}$ such that 
\vspace{-2mm}
\begin{subequations}
    \begin{equation}
        \cos{\theta_i} = 1 - L_i
    \label{eq:cos2distance}
    \vspace{-2mm}
    \end{equation}
    \begin{equation}
        \sin{\theta_i} = \left( 1 - {\cos{\theta_i}}^2 \right)^{\frac{1}{2}} = \left( 2L_i - L_i^2 \right)^{\frac{1}{2}}
        \label{eq:sin2dist}
        \vspace{-1mm}
    \end{equation}
\end{subequations}

Notably, Eq.~\ref{eq:sin2dist} assumes that $\theta_i$ lies in the range $[0, \pi]$ \cite{maric2020inverse2}, and accordingly, $L_i$ lies in the range $[0, 2]$. This is because $\theta_i$ and $\pi - \theta_i$ correspond to the same value of $L_i$ in our formulation, and we only consider the one that falls within $[0, \pi]$. This introduces an additional challenge when the joint angle limit $\theta_i$ is not a subset of $[0, \pi]$, which will be addressed in Sec.~\ref{subsec:joint-limit}.

After attaching points on $\mathcal{F}_i$, we re-parameterize $^{i-1}\mathbf{T}_i$ by substituting Eq.~\ref{eq:cos2distance} and Eq.~\ref{eq:sin2dist} to Eq.~\ref{eq:DH_matrix}:
\vspace{-3mm}
\begin{equation}
    ^{i-1}\mathbf{T}_i = g(L_i; \alpha_{i-1}, a_{i-1}, d_i) =\left[
    \begin{array}{cccc}
        1 - L_i & -{\scriptstyle \left( 2L_i - L_i^2 \right)^{\frac{1}{2}}} & 0 & a \\
        c_\alpha {\scriptstyle \left( 2L_i - L_i^2 \right)^{\frac{1}{2}}} & c_\alpha (1 - L_i) & -s_\alpha & ds_\alpha \\
        s_\alpha {\scriptstyle \left( 2L_i - L_i^2 \right)^{\frac{1}{2}}} & s_\alpha (1 - L_i) & c_\alpha & dc_\alpha \\
        0 & 0 & 0 & 1
    \end{array}
    \right]
    \label{eq:distanceDH_matrix}
\vspace{-2mm}
\end{equation}

The transformation of frame $\mathcal{F}_i$ with respect to the world frame, which is $\mathbf{T}_i$, can be computed by recursively multiplying from $^{0}\mathbf{T}_1$ to $^{i-1}\mathbf{T}_i$. Then, we can extract the position of the joint $\mathbf{u}_i$ from $\mathbf{T}_i$:
\begin{subequations}
\vspace{-3mm}
\begin{multicols}{2}\noindent
\begin{equation}
    \mathbf{T}_i = \prod_{p=1}^{i} {^{p-1}\mathbf{T}_p}
    \label{eq:DHchain}
\vspace{-2mm}
\end{equation}
\begin{equation}
    \mathbf{u}_i = \left[ \mathbf{T}_i^{(1,4)}, \ \mathbf{T}_i^{(2,4)}, \ \mathbf{T}_i^{(3,4)} \right]^\top
    \label{eq:ui_extract}
\vspace{-2mm}
\end{equation}
\end{multicols}
\end{subequations}

\subsection{Joint Limit Constraints}\label{subsec:joint-limit}
Joint limits refer to the restrictions on the range of motion for each joint. This paper focuses on the maximum and minimum joint angle limitations, which are formulated as box constraints $\theta_i \in [ \theta_i^{\text{min}}, \theta_i^{\text{max}} ]$. Our method handles the joint angle constraints by limiting the corresponding squared distance $L_i$ between $L_i^{\text{min}}$ and $L_i^{\text{max}}$. Then, we convert the box constraints to equality constraints by introducing a squashing function based on the sigmoid function \cite{han1995influence}.

The formulation of $L_i$ in Eq.~\ref{eq:sin2dist} assumes that $\theta_i$ is within the range $[0, \pi]$. However, the joint angle limits in real-world robotic arms often exceed this range. We address this issue with angle decomposition. Let $\theta_i^{\rm min} = 0$ and $k = \theta_i^{\text{max}} / \pi > 0$ so that $\theta_i \in [0, k\pi]$.
We divide $\theta_i$ into $\lceil k \rceil$ sub-angles $\theta_{im}$ so that all of the sub-angles lie within the range of $[0, \pi]$:
\begin{equation}
    \begin{array}{lr}
        \theta_i = \sum_{m=1}^{\lceil k \rceil} \theta_{im}, & \quad \theta_{im} \in \left[ \theta_i^{\rm min}/{\lceil k \rceil}, {k\pi}/{\lceil k \rceil} \right] \subseteq [0, \pi]
    \end{array}
    \label{eq:theta_im-limit}
\end{equation}

Fig.~\ref{fig:kinematics-model}b shows an example of angle decomposition when $m=3$. By dividing joint angles into sub-angles, we further attach additional points $\mathbf{w}_{im}$ on the side of $\theta_{im}$, and use $l_m$ to describe the distance between $\mathbf{w}_{im}$ and $\mathbf{w}_{i(m-1)}'$. Again, we take the distance between $\mathbf{w}_{im}$ and $\mathbf{u}_i$ as $1 / \sqrt{2}$. $L_{im}$ is the corresponding squared distance of $\theta_{im}$ such that
\vspace{-1mm}
\begin{equation}
    \cos{\theta_{im}} = 1 - L_{im}
    \label{eq:Lim-limit}
\vspace{-1mm}
\end{equation}

By substituting Eq.~\ref{eq:theta_im-limit} into Eq.~\ref{eq:Lim-limit}, the range of $L_{im}$ is:
\vspace{-1mm}
\begin{equation}
    1 - \cos{(\theta_i^{\rm min}/{\lceil k \rceil})} \leq L_{im} \leq 1 - \cos{({k\pi}/{\lceil k \rceil})}
\vspace{-1mm}
\end{equation}

With the above decomposition process, we can compute $^{i-1}\mathbf{T}_i$ using $L_{im}$:
\vspace{-1mm}
\begin{equation}
    ^{i-1}\mathbf{T}_i = \prod_{m=1}^{{\lceil k \rceil}} {g(L_{im})}
\vspace{-1mm}
\end{equation}

% Notably, $\lceil k \rceil = 1$ when $k \leq 1$. In this case, we can directly project the range of $\theta_i$ to $L_i$:
% \vspace{-2mm}
% \begin{equation}
%     1 - \cos{\theta_i^{\rm min}} \leq L_i \leq 1 - \cos{(k\pi)}
%     \label{eq:Li-limit}
% \vspace{-2mm}
% \end{equation}

So far, we have transformed the joint limit constraint to the box constraints on $L_i$ or $L_{im}$. Box constraints are commonly handled by clamping the variables \cite{khokar2015implementation}, penalizing over constraint violation \cite{tassa2014control,lantoine2008hybrid}, or converting to equality constraints \cite{baumgartner2022fast,marti2020squash}. In this work, we convert the box constraint to equality constraints using a squashing function $s(\omega)$ with the following properties \cite{marti2020squash}:
\vspace{-1mm}
\begin{subequations}
    \begin{equation}
        s(\omega) : \mathbb{R} \rightarrow (s^{\rm min}, s^{\rm max})
    \vspace{-2mm}
    \end{equation}
    \vspace{-2mm}
    \begin{equation}
        \frac{\mathrm{d}}{\mathrm{d}\omega}s(\omega) \geq 0
    \end{equation}
    \vspace{-2mm}
    \begin{equation}
        s^{\rm min} = \lim_{\omega \rightarrow -\infty}s(\omega), \ s^{\rm max} = \lim_{\omega \rightarrow \infty}s(\omega)
    \end{equation}
\end{subequations}
where $\omega \in \mathbb{R}$ is the slack variable. We build our squashing function for $L_i$ upon the sigmoid function \cite{han1995influence}
\vspace{-2mm}
\begin{equation}
    \sigma(\omega_i) = \frac{1}{1 + e^{-\omega_i}} : \mathbb{R} \rightarrow (-1, 1)
    \label{eq:sigmoid}
\vspace{-2mm}
\end{equation}
by linearly scaling $\sigma(\omega_i)$ with $(L_i^{\rm max} - L_i^{\rm min})$ and adding a bias term $L_i^{\rm min}$:
\begin{subequations}
\vspace{-1mm}
    \begin{equation}
         L_i = s(\omega_i) = \left(L_i^{\rm max} - L_i^{\rm min}\right)\sigma(\omega_i) + L_i^{\rm min}
         \label{eq:squash1}
    \vspace{-2mm}
    \end{equation}
    \begin{equation}
        s(\omega_i): \mathbb{R} \rightarrow (L_i^{\rm min}, L_i^{\rm max})
        \label{eq:squash2}
    \vspace{-1mm}
    \end{equation}
\end{subequations}

Eq.~\ref{eq:squash1} and Eq.~\ref{eq:squash2} bounds the squared distance $L_i$ within $(L_i^{\rm min}, L_i^{\rm max})$, which is a close approximation to $[L_i^{\rm min}, L_i^{\rm max}]$. The same constraint conversion can also be applied to $L_{im}$. 

We arrange the slack variables $\omega_i$ or $\omega_{im}$ into a vector $\boldsymbol{\omega} \in \mathbb{R}^{M'}$, where $M' \geq M$ due to the presence of angle decomposition. In our optimization algorithm, which will be detailed in Sec.~\ref{sec:algorithm}, we directly optimize over the slack variable $\boldsymbol{\omega}$. This brings several advantages at the cost of additional non-linearity: First, the squashing function is a smooth function that eliminates issues at box constraint boundaries where the gradient might be zero, discontinuous, or undefined. Moreover, the squashing function avoids the need for explicit checks and enforcement of box constraints during the optimization process.

\vspace{-3mm}
\subsection{Collision Avoidance Constraints}\label{subsec:coll_avoid}
Collision avoidance constraints ensure that the robot does not overlap with obstacles. Distance-based IK methods utilize points attached to the robot and use the minimum distance between these points and obstacles for collision avoidance~\cite{maric2021riemannian}. This models the robot's occupation space as a collection of spheres. Achieving an accurate occupation space requires attaching many points to the robot, making the collision avoidance constraints computationally expensive.

We propose a novel formulation of collision avoidance that only requires the points attached to joints to achieve full-body collision avoidance (Fig.~\ref{fig:kinematics-model}c). Moreover, to better handle unstructured obstacles, our framework considers the obstacles as clusters of points rather than as spheres. This is inspired by the common use of LiDAR or depth cameras in robots, which detect and represent environmental obstacles as point clouds. Consequently, clusters of points, or point clouds, are a natural modality for obstacle representations. We separately consider the collision avoidance formulation for robot joints and links.

The occupation space of the joint is considered as a sphere, whose center is at $\mathbf{u}_i$ and the radius is $r_i$. For joint $i$, we require the distance between $\mathbf{u}_i$ and each obstacles point $\mathbf{o}_j$ no smaller than a minimum distance $r_i$:
\vspace{-2mm}
\begin{equation}
    c_{ij}^{\rm joint} =  r_i - ||\mathbf{u}_i - \mathbf{o}_j|| \leq 0
    \label{eq:joint_coll_avoid}
\vspace{-2mm}
\end{equation}

We consider link $i$ in a serial robot as a straight, thin, and long bar or similar shapes that start from point $\mathbf{u}_i$ and ends at $\mathbf{u}_{i+1}$. As the collision avoidance constraint for link $i$, we enforce the half sum of $||\mathbf{u}_i - \mathbf{o}_j||$ and $||\mathbf{u}_{i+1} - \mathbf{o}_j||$ to be equal or greater than a fixed distance $a_i$:
\vspace{-2mm}
\begin{equation}
    c_{ij}^{\rm link} = 2a_i - (||\mathbf{u}_i - \mathbf{o}_j|| + ||\mathbf{u}_{i+1} - \mathbf{o}_j||) \leq 0
    \label{eq:link_coll_avoid}
\vspace{-2mm}
\end{equation}

Eq.~\ref{eq:link_coll_avoid} indicates that the occupation space of link $i$ is bounded with a prolate spheroid, whose foci are $\mathbf{u}_i$ and $\mathbf{u}_{i+1}$ and semi-major axis is $a_i$. Let $c(\boldsymbol{\omega})$ be the vector containing $c_{ij}^{\rm joint}$ and $c_{ij}^{\rm link}$, the collision avoidance constraint is:
\vspace{-1mm}
\begin{equation}
    c(\boldsymbol{\omega}) \leq 0
\vspace{-2mm}
\end{equation}

\subsection{End Effector Pose Objective}\label{subsec:ee-pose}
The end effector pose objective describes the error between the transformation of the end effector frame and the goal frame. We formulate this objective with the distance between a set of distinct points attached on the end effector frame and the goal frame. We attach $\mathbf{u}_{\rm e}$ on the origin of $\mathcal{F}_{\rm e}$. We then attach a set of distinct points $\mathbf{w}_{{\rm e}p}$ and $\mathbf{v}_{{\rm e}p}$ on $\mathbf{x}_{\rm e}$ and $\mathbf{y}_{\rm e}$, respectively. The position of these points could be computed from $\mathbf{T}_{\rm e}$:
% \vspace{-2mm}
% \begin{subequations}
% \begin{multicols}{3}\noindent
%     \begin{equation}
%         \mathbf{T}_{\rm e} = \left[
%         \begin{array}{cccc}
%             \mathbf{x}_{\rm e} & \mathbf{y}_{\rm e} & \mathbf{x}_{\rm e}\!\times\!\mathbf{y}_{\rm e} & \mathbf{u}_{\rm e} \\
%             0 & 0 & 0 & 1
%         \end{array}
%         \right]
%     \end{equation}
%     \begin{equation}
%         \mathbf{x}_{\rm e} = \frac{\mathbf{w}_{{\rm e}p} - \mathbf{u}_{\rm e}}{k_p}
%         \label{eq:vwep}
%     \end{equation}
%     \begin{equation}
%         \mathbf{y}_{\rm e} = \frac{\mathbf{v}_{{\rm e}p} - \mathbf{u}_{\rm e}}{q_p}
%     \end{equation}
% \end{multicols}
% \end{subequations}
\vspace{-2mm}
\begin{equation}
    \mathbf{w}_{{\rm e}p} = k_p\mathbf{x}_{\rm e} + \mathbf{u}_{\rm e}, \ \mathbf{v}_{{\rm e}p} = q_p\mathbf{y}_{\rm e} + \mathbf{u}_{\rm e}
    \label{eq:vwep}
\vspace{-2mm}
\end{equation}
where $\mathbf{x}_{\rm e}$ and $\mathbf{y}_{\rm e}$ can be extracted from $\mathbf{T}_{\rm e}$. $k_p, q_p \in \mathbb{R}$ and $k_p, q_p \neq 0$. Similarly, we attach $\mathbf{u}^*$, $\mathbf{w}_i^*$, and $\mathbf{v}_i^*$ on the goal frame $\mathcal{F}^*$. Fig.~\ref{fig:kinematics-model}d shows an example when $n_1=n_2=2$.

We arrange the points attached on $\mathcal{F}_{\rm e}$ into matrix $\mathbf{U}_{\rm e} = [\mathbf{u}_{\rm e}, \mathbf{w}_{\rm e1}, ..., \mathbf{w}_{{\rm e}n1}$, $\mathbf{v}_{\rm e1}, ..., \mathbf{v}_{{\rm e}n2}]^\top \in \mathbb{R}^{(n_1+n_2+1)\times3}$ and goal points $\mathbf{u}_p^*$ into matrix $\mathbf{U}^* = [\mathbf{u}^*, \mathbf{w}_{1}^*, ...$, $\mathbf{w}_{n1}^*$, $\mathbf{v}_{1}^*, ..., \mathbf{v}_{n2}^*]^\top \in \mathbb{R}^{(n_1+n_2+1)\times3}$ . The end effector pose objective can be formulated as:
\vspace{-2mm}
\begin{equation}
    J(\boldsymbol{\omega}) = \frac{1}{2} \sum \left( (\mathbf{U}_{\rm e} - \mathbf{U}^*)^\top(\mathbf{U}_{\rm e} - \mathbf{U}^*) \right)
    \label{eq:obj}
\vspace{-1mm}
\end{equation}

Besides the 6-\ac{DOF} end effector pose objective, our method can also handle 5-\ac{DOF} objective by setting $n_2 = 0$ or 3-\ac{DOF} objective by setting $n_1=n_2=0$.

\section{Algorithm}\label{sec:algorithm}
So far, we have unified the formulation of kinematic chain, joint limit constraints, collision avoidance constraints and end effector pose objective into Euclidean distance representations. In this section, we introduce our method of solving the distance-based constrained \ac{IK}. 

We formulate the constrained \ac{IK} as a local optimization problem over the slack variable $\boldsymbol{\omega}$ introduced in Sec.~\ref{subsec:joint-limit}:
\vspace{-2mm}
\begin{equation}
    \begin{aligned}
        \min_{\boldsymbol{\omega}} \quad & J(\boldsymbol{\omega}) = \frac{1}{2} \sum  (\mathbf{U}_{\rm e} - \mathbf{U}^*)^\top(\mathbf{U}_{\rm e} - \mathbf{U}^*)  \\
        \textrm{s.t.} \quad & \mathbf{U}_{\rm e} = FK(\boldsymbol{\omega}) \\
        & c(\boldsymbol{\omega}) \leq 0
    \end{aligned}
    \label{eq:local-opt-form}
\vspace{-3mm}
\end{equation}
where $FK$ is the forward kinematics computation.

The equality constraint, projecting from $\boldsymbol{\omega}$ to the end effector pose via the kinematic chain, can be directly incorporated into $J(\boldsymbol{\omega})$ by replacing $\mathbf{U}{\text{e}}$ with ${FK}(\boldsymbol{\omega})$. Then, we convert the inequality constraint into penalty terms to form an augmented Lagrangian function $L{\rho}$. Let $c'(\boldsymbol{\omega}) = \max(0, c(\boldsymbol{\omega}))$, the augmented Lagrangian function is:
\vspace{-2mm}
\begin{equation}
    L_\rho(\boldsymbol{\omega}) = J(\boldsymbol{\omega}) + \boldsymbol{\mu}^\top c'(\boldsymbol{\omega}) + \frac{\rho}{2}  c'(\boldsymbol{\omega})^\top c'(\boldsymbol{\omega})
    \label{eq:AL}
\vspace{-2mm}
\end{equation}
where $\lambda$ is the Lagrangian multiplier and $\rho$ is the adjust penalty parameter. The constrained \ac{IK} problem is formulated as finding $\boldsymbol{\omega}^* \in \mathbb{R}^{M'}$ such that
\vspace{-2mm}
\begin{equation}
    \boldsymbol{\omega}^* = {\rm argmin} L_\rho(\boldsymbol{\omega})
\vspace{-2mm}
\end{equation}

\subsection{Forward Rollout}\label{subsec:FWD-Algorithm}
The forward rollout procedure involves the computation of the forward kinematics $FK$ and objective $L_\rho$. The forward kinematics computes the position of points attached on joints $\mathbf{U}$ given motion variable $\boldsymbol{\omega}$. 

As shown in Fig.~\ref{fig:kinematics-model}e and Algorithm~\ref{alg:FWD}, the forward rollout is computed propagatively along the direction of the kinematic chain from the base to the end effector. Fig.~\ref{fig:kinematics-model}e illustrates a diagonal matrix of distances between points attached on robot. Each element represents the distance between points at the row and column indices. The propagation starts from $\mathbf{u}_0$, $\mathbf{w}_0$, and $\mathbf{v}_0$, which are assumed to be pre-known since they are typically stationary relative to the world frame. We can collect the set of $\mathbf{u}_i$, $\mathbf{w}_i$, and $\mathbf{v}_i$ as a \textit{unit}, then the forward rollout solves for the $i$th unit and then moves to the $(i+1)$th unit. Additionally, the computation of the variables in the $i$th unit reuses the pre-computed variables in the $(i-1)$th unit.

After \ac{FK} computation, we compute the augmented Lagragian $L_\rho$, which is composed of end effector pose objective $J(\boldsymbol{\omega})$ and collision penalty $c(\boldsymbol{\omega})$. $J(\boldsymbol{\omega})$ is computed with $\mathbf{U}_{\rm e}(\boldsymbol{\omega})$ obtained from \ac{FK}. For collision penalties, we loop through every robot-attached point $\mathbf{u}_i$ and obstacle point $\mathbf{o}_j$ to compute $c_{ij}^{\rm joint}$ and $c_{ij}^{\rm link}$.

{\footnotesize
\begin{algorithm}[t]
\caption{Forward Rollout}\label{alg:FWD}
    \vspace{-5mm}
    \begin{multicols}{2}
    \begin{algorithmic}[1]
    \State{$\mathbf{T}_0 \leftarrow \mathbf{T}_{\rm world}$}
    \For{$i = 1, 2, ..., M$}%\Comment{Forward kinematics}
        \State{$L_i \leftarrow s(\omega_i)$}\Comment{Eq.~\ref{eq:squash1}}
        \State{$^{i-1}\mathbf{T}_i \leftarrow g(L_i)$}\Comment{Eq.~\ref{eq:distanceDH_matrix}}
        \State{$\mathbf{T}_i \leftarrow \mathbf{T}_{i-1} \cdot {^{i-1}\mathbf{T}_i}$}\Comment{Eq.~\ref{eq:DHchain}}
        \State{Compute $\mathbf{u}_i$ from $\mathbf{T}_i$}\Comment{Eq.~\ref{eq:ui_extract}}
    \EndFor
    \State{$\mathbf{u}_{\rm e} \leftarrow \mathbf{u}_M$}%\Comment{End effector pose objective}
    \State{Compute $\mathbf{w}_{{\rm e}p1}$ and $\mathbf{v}_{{\rm e}p2}$ for $p_1 = 1, ..., n_1$ and $p_2 = 1, ..., n_2$.}
    \State{$\mathbf{U}_{\rm e} \leftarrow [\mathbf{u}_{\rm e}, \mathbf{u}_{\rm e1}, ..., \mathbf{u}_{{\rm e}n}]^\top$.}\Comment{Eq.~\ref{eq:vwep}}
    \State{Compute $J$ from $\mathbf{U}_{\rm e}$ and $\mathbf{U}^*$.}\Comment{Eq.~\ref{eq:obj}}
    \State{$L_\rho \leftarrow J$}
    \For{$j = 1, 2, ..., N$}%\Comment{Collision penalty}
        \For{$i = 1, 2, ..., M$}
            \State{$c_{ij}^{\rm joint} \leftarrow r_i - ||\mathbf{u}_i - \mathbf{o}_j||$}\Comment{Eq.~\ref{eq:joint_coll_avoid}}
            \State{$c_{ij}^{\rm link} \leftarrow 2a_i - (||\mathbf{u}_i - \mathbf{o}_j|| + ||\mathbf{u}_{i+1} - \mathbf{o}_j||)$}\Comment{Eq.~\ref{eq:link_coll_avoid}}
            \State{$L_\rho \leftarrow L_\rho + \mu_{ij}c_{ij}^{\rm joint} + \mu_{ij}c_{ij}^{\rm link} + \frac{\rho}{2}{c_{ij}^{\rm joint}}^2 + \frac{\rho}{2}{c_{ij}^{\rm link}}^2$}\Comment{Eq.~\ref{eq:AL}}
        \EndFor
    \EndFor
    \State{\textbf{return} $L_\rho$}
    \end{algorithmic}
    \end{multicols}
    \vspace{-3.5mm}
\end{algorithm}
}

\vspace{-3mm}
\subsection{Jacobian Computation}\label{subsec:jacobian}
The Jacobian $\nabla_{\boldsymbol{\omega}}L_\rho$ is the derivatives of augmented Lagrangian $L_\rho$ to the variables $\boldsymbol{\omega}$, which can be decomposed into the derivatives of end effector pose objective $\nabla_{\boldsymbol{\omega}}J(\boldsymbol{\omega})$ and collision penalties $\nabla_{\boldsymbol{\omega}}c(\boldsymbol{\omega})$:
\vspace{-2mm}
\begin{subequations}
    \begin{equation}
        \nabla_{\boldsymbol{\omega}}L_\rho = \nabla_{\boldsymbol{\omega}}J(\boldsymbol{\omega}) + \boldsymbol{\mu} \nabla_{\boldsymbol{\omega}}c(\boldsymbol{\omega}) + \rho  c(\boldsymbol{\omega}) \odot \nabla_{\boldsymbol{\omega}}c(\boldsymbol{\omega})
    \vspace{-2mm}
    \end{equation}
    \begin{multicols}{2}\noindent
    \begin{equation}
        \nabla_{\boldsymbol{\omega}}J(\boldsymbol{\omega}) = \frac{\mathrm{d}J}{\mathrm{d}\mathbf{T}_{\rm e}}\frac{\mathrm{d}\mathbf{T}_{\rm e}}{\mathrm{d}\boldsymbol{\omega}}
        \label{eq:dJdomega}
    \end{equation}
    \begin{equation}
        \nabla_{\boldsymbol{\omega}}c(\boldsymbol{\omega}) = \sum_{i=1}^M\frac{\partial c}{\partial \mathbf{u}_i}\frac{\mathrm{d} \mathbf{u}_i}{\mathrm{d}\boldsymbol{\omega}}
        \label{eq:dcdomega}
    \end{equation}
    \end{multicols}
\vspace{-3mm}
\end{subequations}

{\footnotesize
\begin{algorithm}[t]
	\caption{Jacobian Computation}\label{alg:Jacobian}
	\begin{algorithmic}[1]
        \For{$j = N, N-1, ..., 1$}%\Comment{Collision penalties}
            \For{$i = M, M-1,..., 1$}
                \If{$c_{ij}^{\rm joint} > 0$}
        		  \State{$\mathbf{s}_1 \leftarrow (\mu_{ij} + \rho c_{ij}^{\rm joint})(\mathbf{u}_i - \mathbf{o}_j) / ||\mathbf{u}_i - \mathbf{o}_j||$}\Comment{Eq.~\ref{eq:dcdu-joint}}
                \State{Add $[\mathbf{s}_1^\top, 0]^\top$ to the last column of $\partial L_\rho / \partial \mathbf{T}_i$.}\Comment{Eq.~\ref{eq:dcdu}}
                \EndIf
                \If{$c_{ij}^{\rm link} > 0$}
        		  \State{$\mathbf{s}_2 \leftarrow (\mu_{ij} + \rho c_{ij}^{\rm link})(\mathbf{u}_i - \mathbf{o}_j) / ||\mathbf{u}_i - \mathbf{o}_j||$}
                \State{Add $[\mathbf{s}_2^\top, 0]^\top$ to the last column of $\partial L_\rho / \partial \mathbf{T}_i$.}\Comment{Eq.~\ref{eq:dcdu-link}}
                \State{$\mathbf{s}_3 \leftarrow (\mu_{(ij} + \rho c_{(i-1)j}^{\rm link})(\mathbf{u}_{i-1} - \mathbf{o}_j) / ||\mathbf{u}_{i-1} - \mathbf{o}_j||$}\Comment{Eq.~\ref{eq:dcdu-link}}
                \State{Add $[\mathbf{s}_3^\top, 0]^\top$ to the last column of $\partial L_\rho / \partial \mathbf{T}_{i-1}$.}\Comment{Eq.~\ref{eq:dcdu}}
        		\EndIf
            \EndFor
        \EndFor
        \State{$\partial L_\rho / \partial \mathbf{T}_M \leftarrow \partial J / \partial \mathbf{T}_{\rm e}$}\Comment{Eq.~\ref{eq:dJdTe}}
		\For{$i = M, M-1, ..., 1$}%\Comment{Derivatives backward propagation}
            \State{$\partial L_\rho / \partial {^{i-1}\mathbf{T}_i} \leftarrow \mathbf{T}_{i-1}^\top \cdot (\partial L_\rho / \partial \mathbf{T}_{i})$}\Comment{Eq.~\ref{eq:dLdDH}}
            \State{$\partial {^{i-1}\mathbf{T}_i} / \partial L_i \leftarrow g'(L_i)$}\Comment{Eq.~\ref{eq:dDHddistsquare}}
            \State{$\partial L_i / \partial \omega_i \leftarrow (L_i^{\rm max} - L_i^{\rm min})\sigma(\omega_i)(1-\sigma(\omega_i))$}\Comment{Eq.~\ref{eq:ddistsquaredomega}}
            \State{$\partial L_\rho / \partial \omega_i \leftarrow \left(\sum(\partial L / \partial {^{i-1}\mathbf{T}_i}) \cdot (\partial {^{i-1}\mathbf{T}_i} / \partial L_i)\right)(\partial L_i / \partial \omega_i)$}\Comment{Eq.~\ref{eq:dLdw}}
            \State{$\partial L_\rho / \partial \mathbf{T}_{i-1} \leftarrow \partial L_\rho / \partial \mathbf{T}_{i-1} + (\partial L_\rho / \partial \mathbf{T}_{i}) \cdot ^{i-1}\mathbf{T}_i^\top$}\Comment{Eq.~\ref{eq:dLdT}}
		\EndFor
        \State{$\nabla_{\boldsymbol{\omega}}L_\rho \leftarrow [\partial L_\rho / \partial \omega_1, \partial L_\rho / \partial \omega_2, ..., \partial L_\rho / \partial \omega_i, ..., \partial L_\rho / \partial \omega_M]^\top$}
		\State{\textbf{return} $\nabla_{\boldsymbol{\omega}}L_\rho$}
	\end{algorithmic}
\vspace{-1mm}
\end{algorithm}
}

The key idea of Jacobian computation in our method is to propagate along the backward direction of the kinematic chain leveraging reverse accumulation. Notably, the Jacobian computation follows the forward rollout, allowing it to reuse the results from the forward rollout. As shown in Algorithm~\ref{alg:Jacobian}, we first compute the derivatives of collision penalties with respect to the position of points attached to the robot (${\partial c} / {\partial \mathbf{u}i}$). Then, we compute the derivative of the end effector pose objective to the end frame (${\partial J} / {\partial \mathbf{T}{\text{e}}}$), and finally compute the derivatives of the position of the points to the variables (${\partial \mathbf{u}_i} / {\partial \boldsymbol{\omega}}$). $\partial c / \partial \mathbf{u}_i$ in Eq.~\ref{eq:dcdomega} is computed with
\vspace{-2mm}
\begin{subequations}
\begin{equation}
    \frac{\partial c}{\partial \mathbf{u}_i} = \sum_{j=1}^N \left(\frac{d}{d\mathbf{u}_i}c_{ij}^{\rm joint} + \frac{\partial}{\partial \mathbf{u}_i}c_{ij}^{\rm link} + \frac{\partial}{\partial \mathbf{u}_i}c_{(i+1)j}^{\rm link} \right)
    \label{eq:dcdu}
\end{equation}
\vspace{-2mm}
\begin{equation}
    \frac{\mathrm{d}}{\mathrm{d}\mathbf{u}_i}c_{ij}^{\rm joint} = (\mu_{ij} + \rho c_{ij}^{\rm joint})\frac{\mathbf{u}_i - \mathbf{o}_j}{||\mathbf{u}_i - \mathbf{o}_j||} \cdot \mathds{1}\left(c_{ij}^{\rm joint} \geq 0\right)
    \label{eq:dcdu-joint}
\vspace{-1mm}
\end{equation}
\begin{equation}
    \frac{\mathrm{d}}{\mathrm{d}\mathbf{u}_i}c_{ij}^{\rm link} = (\mu_{ij} + \rho c_{ij}^{\rm link})\frac{\mathbf{u}_i - \mathbf{o}_j}{||\mathbf{u}_i - \mathbf{o}_j||} \cdot \mathds{1}\left(c_{ij}^{\rm link} \geq 0\right)
    \label{eq:dcdu-link}
\end{equation}
\end{subequations}
where the term $\partial c_{(i+1)j}^{\rm link} / \partial \mathbf{u}_i$ is dropped when $i = M$. The term $\mathbf{u}_i - \mathbf{o}_j$ and $||\mathbf{u}_i - \mathbf{o}_j||$ are already pre-computed in the forward rollout by Eq.~\ref{eq:joint_coll_avoid}. The term $\mathds{1}(\cdot)$ is the indicator function, which equals to 1 when $(\cdot)$ is true otherwise 0. The term $\partial J / \partial \mathbf{T}_{\rm e}$ in Eq.~\ref{eq:dJdomega} is computed with:
\vspace{-2mm}
\begin{equation}
    \frac{\partial}{\partial \mathbf{T}_{\rm e}}J = \left[
        \begin{array}{cccc}
            \sum_{p=1}^{n_1}k_p(\mathbf{w}_{{\rm e}p} - \mathbf{w}^*) \ & \ \sum_{p=1}^{n_2}q_p(\mathbf{v}_{{\rm e}p} - \mathbf{v}^*) \ & \ 0 \ & \ \mathbf{u}_{\rm e} - \mathbf{u}^* \\
            0 & 0 & 0 & 1
        \end{array}
        \right]
    \label{eq:dJdTe}
\vspace{-2mm}
\end{equation}
where all the elements are already computed in Eq.~\ref{eq:vwep} and Eq.~\ref{eq:obj}. By computing the derivatives of $L_\rho$ to the position of points attached on the robot, we can now solve for $\partial L_\rho / \partial \omega_i$ with:
\vspace{-2mm}
\begin{subequations}
\begin{equation}
    \frac{\partial}{\partial \omega_i}L_\rho = \left(\sum\frac{\partial L_\rho}{\partial {^{i-1}\mathbf{T}_i}}\cdot \frac{\partial {^{i-1}\mathbf{T}_i}}{\partial L_i} \right)\frac{\partial L_i}{\partial \omega_i}
    \label{eq:dLdw}
\end{equation}    
\vspace{-2mm}
\begin{equation}
    \frac{\partial L_\rho}{\partial {^{i-1}\mathbf{T}_i}} = \mathbf{T}_{i-1}^\top \cdot \frac{\partial L_\rho}{\partial \mathbf{T}_{i}}
    \label{eq:dLdDH}
\vspace{-2mm}
\end{equation}
\begin{equation}
{\scriptsize 
    \frac{\partial {^{i-1}\mathbf{T}_i}}{\partial L_i} = \frac{\mathrm{d}g(L_i)}{L_i} = \left[
    \begin{array}{cccc}
        -1 &  -\left(1 - L_i\right)\left( 2L_i - L_i^2 \right)^{-\frac{1}{2}} & 0 & 0 \\
        c_\alpha \left(1 - L_i\right)\left( 2L_i - L_i^2 \right)^{-\frac{1}{2}} & -c_\alpha & 0 & 0 \\
        s_\alpha \left(1 - L_i\right)\left( 2L_i - L_i^2 \right)^{-\frac{1}{2}} & -s_\alpha & 0 & 0 \\
        0 & 0 & 0 & 0
    \end{array}
    \right]
    \label{eq:dDHddistsquare}}
\end{equation}
\vspace{-2mm}
\begin{equation}
    \frac{\partial L_i}{\partial \omega_i} = (L_i^{\rm max} - L_i^{\rm min})\frac{\mathrm{d}}{\mathrm{d}\omega_i}\sigma(\omega_i) = (L_i^{\rm max} - L_i^{\rm min})\sigma(\omega_i)(1-\sigma(\omega_i))
    \label{eq:ddistsquaredomega}
\vspace{-1mm}
\end{equation}
\end{subequations}
where $\mathbf{T}_{i-1}$ is already computed in Eq.~\ref{eq:DHchain}, $\sigma(\omega_i)$ is already computed in Eq.~\ref{eq:sigmoid}, and $\partial L_\rho / \partial \mathbf{T}_{i}$ can be recursively computed from joint $i+1$, collision penalty, and end effector pose objectives:
\vspace{-2mm}
\begin{equation}
    \frac{\partial L_\rho}{\partial \mathbf{T}_{i}} = \left\{
    \begin{array}{cc}
        \frac{\partial c}{\partial \mathbf{T}_{i}} + \frac{\partial L_\rho}{\partial \mathbf{T}_{i+1}} \cdot ^{i}\mathbf{T}_{i+1}^\top &, \ i<M  \\
        \frac{\partial c}{\partial \mathbf{T}_{i}} + \frac{\partial J}{\partial \mathbf{T}_{\rm e}} &, \ i=M 
    \end{array}
    \right.
    \label{eq:dLdT}
\vspace{-2mm}
\end{equation}

\vspace{-3mm}
\subsection{Inverse Kinematics}\label{subsec:IK-Algorithm}
We first solve Eq.~\ref{eq:AL} for a local optimal solution $\boldsymbol{\omega}^*$, and then compute $\boldsymbol{\theta}^*$ from $\boldsymbol{\omega}^*$. As shown in Algorithm~\ref{alg:IK}, our method iteratively minimize $L_\rho$ and update the Lagrangian multiplier $\boldsymbol{\mu}$ and $\rho$. Within each loop $k$, we use quasi-Newton method to solve for ${\rm argmin}L_\rho$. The forward rollout and Jacobian computation are implemented following Algorithm~\ref{alg:FWD} and Algorithm~\ref{alg:Jacobian}, respectively. The Hessian matrix are approximated with the Jacobian in quasi-Newton convention. In our framework, we use \ac{L-BFGS} as our quasi-Newton-based solver. After solving for $\boldsymbol{\omega}_k^* = {\rm argmin}L_\rho$, we update Lagrangian multipliers $\boldsymbol{\mu}$ and $\rho$. Our method checks ${\rm max}(c(\boldsymbol{\omega}_k^*))$ in every iteration and will terminate if ${\rm max}(c(\boldsymbol{\omega}_k^*)) < c_{\rm tol}$ \textbf{or} ${\rm max}(c(\boldsymbol{\omega}_k^*)) \geq \beta c_{\rm last}$, where $\beta < 1$ and $c_{\rm last}={\rm max}(c(\boldsymbol{\omega}_{k-1}^*))$.

{\footnotesize
\begin{algorithm}[t]
	\caption{Inverse Kinematics}\label{alg:IK}
         \vspace{-5mm}
        \begin{multicols}{2}
        \begin{algorithmic}[1]
		\State{$\boldsymbol{\omega}_0^* \leftarrow \mathbf{0}$, $\boldsymbol{\mu} \leftarrow \mathbf{0}$, $\rho \leftarrow 1$, $\alpha \leftarrow 10$, $c_{\rm last} \leftarrow \infty$.}
		\For{iteration $k = 1, 2, ..., k_{\rm max}$}
        \State{$\boldsymbol{\omega}_k^* \leftarrow L-BFGS(\boldsymbol{\omega}_{k-1}^*)$, where $L_\rho$ is computed with Algorithm~\ref{alg:FWD} and $\nabla_{\boldsymbol{\omega}_k} L_\rho$ is computed with Algorithm~\ref{alg:Jacobian}.}
        \If{${\rm max}(c(\boldsymbol{\omega}_k^*)) < c_{\rm tol}$ \textbf{or} ${\rm max}(c(\boldsymbol{\omega}_k^*)) \geq \beta c_{\rm last}$}
        \State{$\boldsymbol{\omega}^* \leftarrow \boldsymbol{\omega}_k^*$}
        \State{\textbf{break}}
        \EndIf
        \State{$c_{\rm last} \leftarrow {\rm max}(c(\boldsymbol{\omega}_k^*))$}
        \State{$\boldsymbol{\mu} \leftarrow \boldsymbol{\mu} + \rho c(\boldsymbol{\omega}_k^*)$}
        \State{$\rho \leftarrow \alpha\rho$}
		\EndFor
        \State{Compute $\boldsymbol{\theta}^*$ with $\boldsymbol{\omega}^*$}\Comment{Eq.~\ref{eq:squash1} and Eq.~\ref{eq:recovery}}
		\State{\textbf{return} $\boldsymbol{\theta}^*$}
	\end{algorithmic}
        \end{multicols}
	\vspace{-3.5mm}
\end{algorithm}
}

By obtaining $\boldsymbol{\omega}^*$, we solve $\boldsymbol{\theta}^*$ from $\boldsymbol{\omega}^*$. We first compute $L_i^*$ using Eq.~\ref{eq:squash1}, and then compute $\theta_i^*$ from $L_i^*$:
\vspace{-3mm}
\begin{equation}
    \theta_i^* = \sum_{m=1}^{\lceil k \rceil} \arccos{(1 - L_{im}^*)} + \min{(0, \theta_i^{\rm min})}
\vspace{-2mm}
\label{eq:recovery}
\end{equation}

\subsection{Complexity Analysis}
\vspace{-2mm}
Our optimization framework is a combination of augmented Lagrangian optimization and a quasi-Newton optimizer. It's complexity is determined by the complexity of a single iteration, which involves forward rollout (Algorithm~\ref{alg:FWD}), Jacobian computation (Algorithm~\ref{alg:Jacobian}), Hessian approximation, and variable update. The forward rollout and Jacobian computation propagate through all of the points attached on the robot and the points of the obstacles, with a complexity of $O(M'N)$. The complexity of Hessian approximation process in quasi-Newton solver is $O(M'^2)$. The complexity of variable update is $O(M')$. In summary, the complexity of the algorithm is $O(M'N + M'^2)$. Notably, $M'$ is only determined by the \ac{DOF} of the robot and joint limit, which is commonly small (less than 20). On the other hand, $N$ depends on the environment and could reach hundreds. Given a robot, the complexity of our algorithm is linear to the number of cluttered points in the environment.

\section{Experiments}\label{sec:exp}
In this section, we conduct experiments to demonstrate the efficiency, effectiveness, and generalization capability of our algorithm.

\vspace{-3mm}
\subsection{Efficiency and Effectiveness Comparison}\label{subsec:exp1}
We conduct simulation experiments to benchmark the efficiency of our algorithm in runtime, as well as its effectiveness in handling end effector pose objective, collision avoidance constraints, and joint limit constraints.

\begin{figure}[t]
    \centering
    \includegraphics[width=0.9\linewidth]{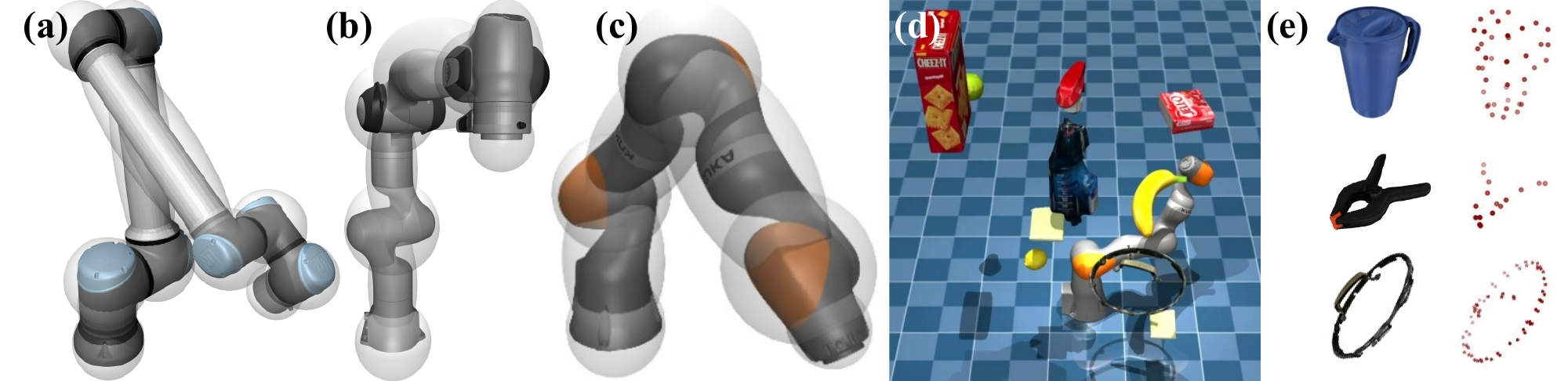}
    \caption{Visualization of experimental setups. (a)-(c) Robot arm platforms (UR10, Franka, and KUKA). Their occupation space are composed of spheres or spheroids in our formulation, visualized here as translucent hulls. (d) An example of a KUKA robot in the environment with 9 random obstacles. (e) Visualization of obstacles as point clusters.}
    \label{fig:exp-setup}
\vspace{-2mm}
\end{figure}

We compare our method with three baselines, two of which are recent distance-based \ac{IK} algorithms. The first baseline is \ac{R-TR} \cite{maric2021riemannian}, a distance-geometric-based IK algorithm that optimizes the distance matrix on the Riemannian manifold using the Trust Region method. The second baseline is \ac{R-CG}, also introduced in \cite{maric2021riemannian}, which uses the Conjugate Gradient method to solve for the distance matrix. Since the baselines can only handle 3- or 5-\ac{DOF} end effector poses, we focus on 5-\ac{DOF} end effector pose objectives throughout the experiments. We further build a third baseline: a variant of \ac{PDO-IK}, which employs the same collision avoidance constraints, end effector pose objectives, and optimization method as \ac{PDO-IK}, but represents robot kinematics using joint angles and uses \ac{L-BFGS-B} \cite{zhu1997algorithm} to directly handle joint limits as box constraints. We name this third baseline ``Angle-LBFGS-B''. For both \ac{PDO-IK} and Angle-LBFGS-B, we set $\beta=0.99$.

We test all the methods on 3 popular commercial robot arms: UR10, KUKA-IIWA, and Franka \footnote{UR10: \url{https://www.universal-robots.com/}. KUKA-IIWA: \url{https://www.kuka.com/}. Franka: \url{https://franka.de/}.}. UR10 has 6 \ac{DOF}s and every joint can rotate from $-360^\circ$ to $360^\circ$. KUKA-IIWA and Franka have 7 \ac{DOF}s and have tighter joint limits.
As mentioned in Sec.~\ref{subsec:coll_avoid}, our collision constraint formulation takes the occupation space of joints and links as sphere and spheroid, respectively. The occupation space of these robots is shown in Fig.\ref{fig:exp-setup}a-c.

\begin{figure}[t]
    \centering
    \includegraphics[width=1\linewidth]{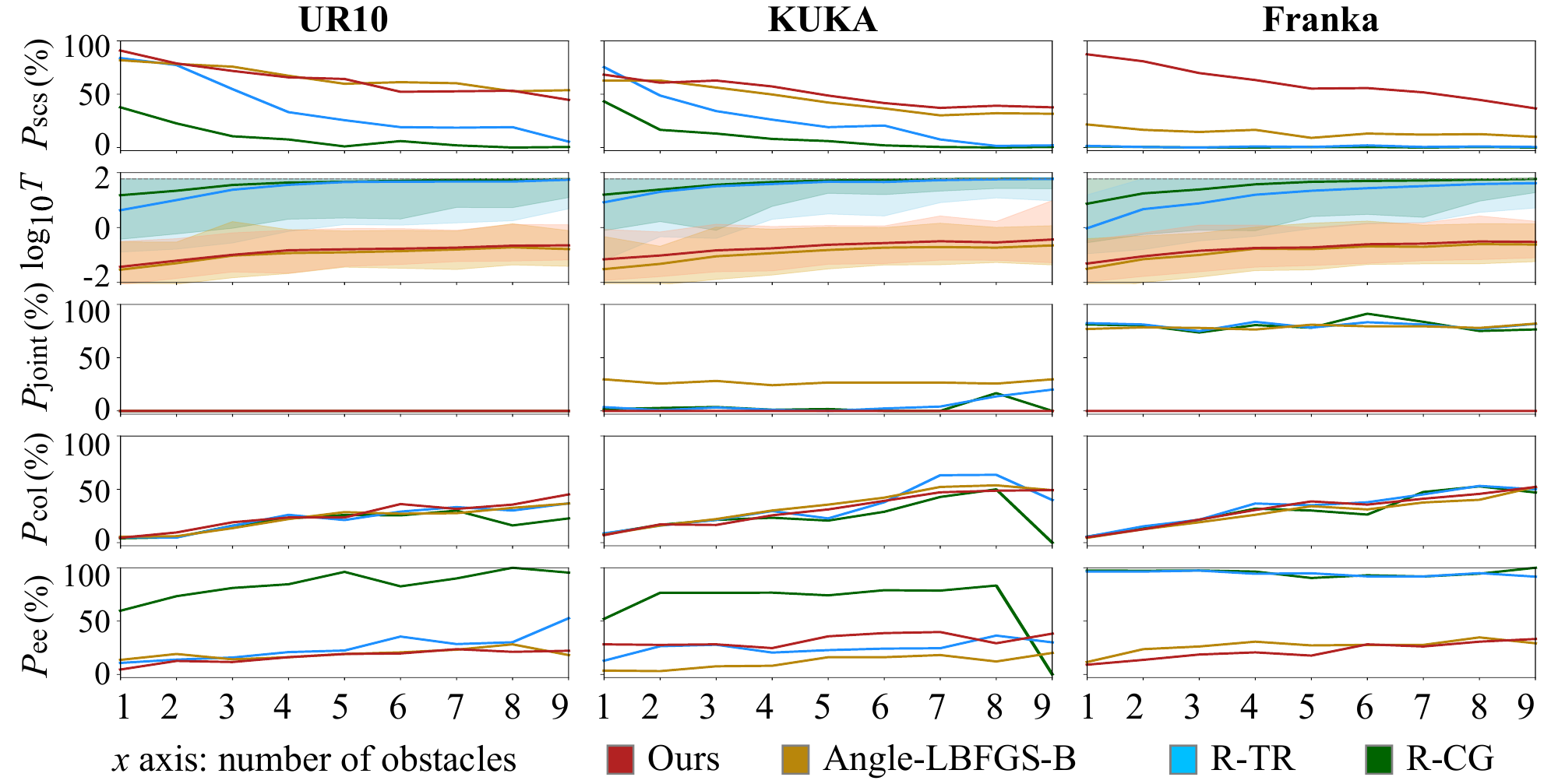}
    \caption{Experimental results on UR10, KUKA, and Franka. The $x$-axis are the number of obstacles. The $y$-axis are success rate, logarithm of runtime in seconds, joint limit violation rate, collision rate, and end effector objective failure rate.}
    \label{fig:exp-result}
\vspace{-5mm}
\end{figure}

We set up various scenarios with 1 to 9 random obstacles. For each number of obstacles and each robot, we generate 200 scenarios for experiments. 
We sample different objects form YCB dataset as obstacles~\cite{calli2015ycb}, with their corresponding point clouds also provided.
% The obstacles and their point clouds are sampled from the YCB dataset \cite{calli2015ycb}. 
An example of an experiment scenario is shown in Fig.\ref{fig:exp-setup}d. The environment generation and method implementation for each scenario follow these steps:
\begin{enumerate}
    \item Place the robot on the origin of the world frame, randomly sample a configuration $\boldsymbol{\theta}_{\rm rand}$ from joint space $\mathcal{C}$ from a uniform distribution over the joint angle limits. Record the end effector pose as the target.
    \item Randomly generate obstacles that are collision free to $\boldsymbol{\theta}_{\rm rand}$. The position of the obstacles are randomly sampled from a uniform distribution within a specific range. The range of the $x$-, $y$-, and $z$-coordinates of the center of the objects are $[-0.6, 0.6]$, $[-0.6, 0.6]$, and $[0, 1.2]$ meters, respectively. The objects are randomly scaled with ratio of $[1, 3]$. We use Moveit! collision checker \footnote{Moveit!: \url{https://moveit.ros.org/}.} to check for collision. The point cloud of the obstacle is downsampled with the Voxel Grid filter from the Point Cloud Library (PCL) \footnote{Point Cloud Library: \url{https://pointclouds.org/}.} with voxel grid leaf size equals to 0.1m. Examples of obstacles and their point cloud are shown in Fig.~\ref{fig:exp-setup}e. The average amount of points in scenarios of 1 to 9 obstacles are 24, 47, 68, 87, 101, 110, 122, 130, and 136, respectively.
    \item Run each \ac{IK} method given the target end effector pose and clusters of points. We use $\boldsymbol{\theta}_0 = \mathbf{0}$ as initialization. We set time limits of 60 seconds for each algorithm. All IK algorithms are implemented in Python on a desktop computer with Intel Core i9 CPU with 128 GB RAM.
\end{enumerate}

We report the success rate $P_{\rm scs}$, which is the is the percentage of experiments that satisfies the following criteria: (1) The solution is reported within the time limit. (2) The solution is collision-free to the obstacles detected by Moveit! collision checker. (3) The end effector position error $\epsilon_{d}$ and rotation error $\epsilon_{\theta}$ are less than 0.01 m and 0.01 rad, respectively. (4) The joint limit violation is within $1\%$ of the joint angle limit range. In addition, we also report joint limit failure rate $P_{\rm joint}$, collision rate $P_{\rm col}$, and end effector pose failure rate $P_{\rm ee}$, which are the proportion of the number of solutions that fails to satisfy criteria (2), (3), and (4), respectively, to the number of total solutions generated.
The logarithm of runtime $\log_{10}T$ where $T$ is in seconds, is also reported for all \ac{IK} methods.

The experiment results are shown in Fig.\ref{fig:exp-result}. For all scenarios and robot platforms, our method achieves a comparable or higher success rate than the baselines, especially when the amount of obstacles in the environment gets higher. Moreover, our method runs up to two orders of magnitude faster than the previous distance-based methods. Our method also achieves lower or comparable $P_{\rm ee}$ compared to previous distance-based methods. The collision rate $P_{\rm col}$ of our method and the baselines are similar. Our method has 0 joint limit failure rate since our joint angle is strictly bounded with the squashing function. Although Angle-LBFGS-B has comparable or slightly faster speed than PDO-IK due to their similar approaches, it can only achieve a comparable success rate to PDO-IK on the UR10, which has 6 DoFs, with wide joint limits, and a simple kinematic chain. On robots with 7 DoFs with tight joint limits and more complex kinematic chains, such as the Franka, Angle-LBFGS-B performs much worse than PDO-IK. This comparison shows the advantage of distance-based representation over conventional angle-based representation.

\begin{figure}[t]
    \centering
    \includegraphics[width=0.9\linewidth]{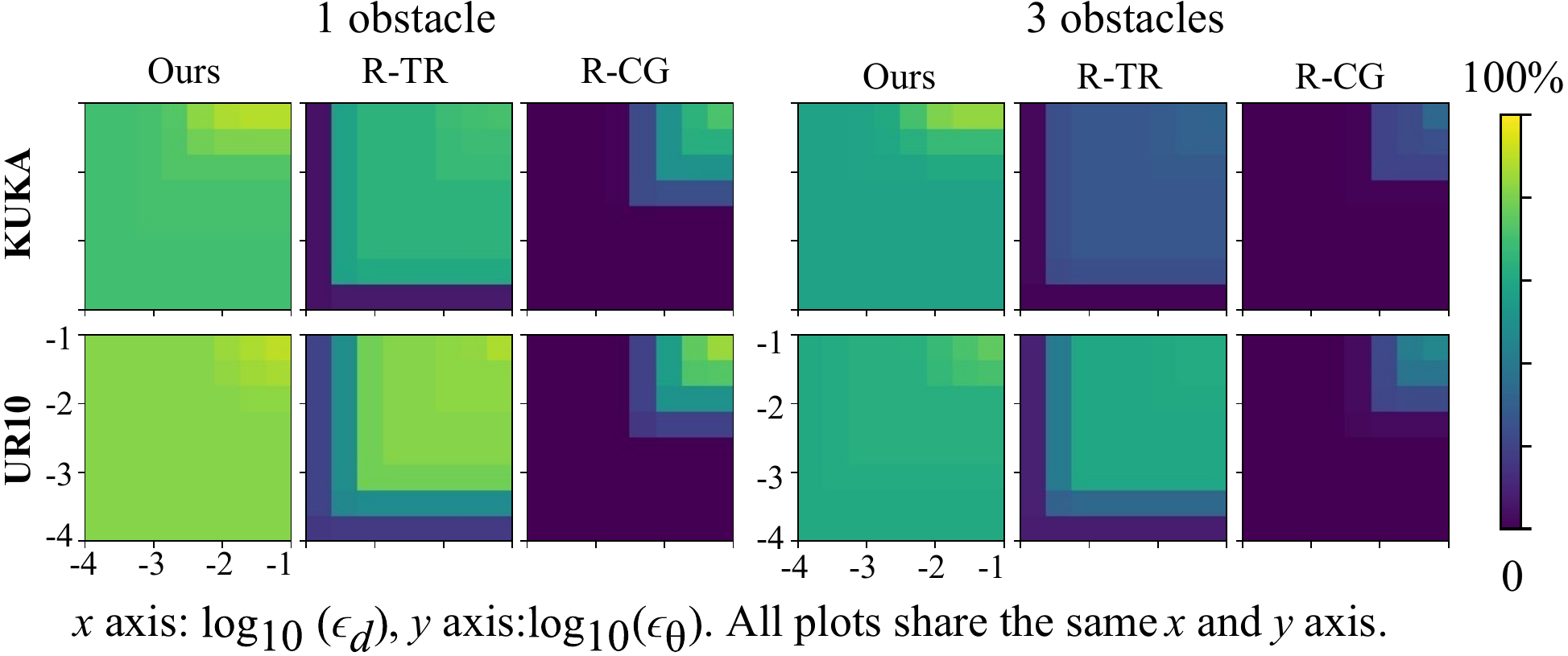}
    \caption{Convergence precision experiments results.}
    \label{fig:exp-acc}
\vspace{-5mm}
\end{figure}

\vspace{-3mm}
\subsection{Solution Accuracy Comparison}\label{subsec:exp2}
We check the solution accuracy with the optimal solution achieved among \ac{PDO-IK}, \ac{R-TR}, and \ac{R-CG}. The solution accuracy measures how close the algorithm's final solution is to the true optimal solution. In this section, we compare the solution accuracy of \ac{PDO-IK}, \ac{R-TR}, and \ac{R-CG} by counting the proportion of end effector objectives that satisfies different tolerance levels of $\epsilon_{d}$ and $\epsilon_{\theta}$.

Fig.~\ref{fig:exp-acc} illustrates the success rate under different tolerance levels of $\epsilon_{d}$ and $\epsilon_{\theta}$ on KUKA and UR10 when the number of obstacles is 1 and 3. Our method remains a high and relative consistent success rate when the tolerance varies from $10^{-4}$ to $10^{-1}$. The success rate of \ac{R-TR} and \ac{R-CG}, however, significantly drop when the tolerance is is below $10^{-3}$ and $10^{-2}$, respectively. This experiment shows that \ac{PDO-IK} achieves higher solution accuracy than the baselines.

\vspace{-3mm}
\subsection{Humanoid Robot Avoiding Dynamic Obstacle}\label{subsec:exp3}
\begin{figure}[t]
    \centering
    \includegraphics[width=1\linewidth]{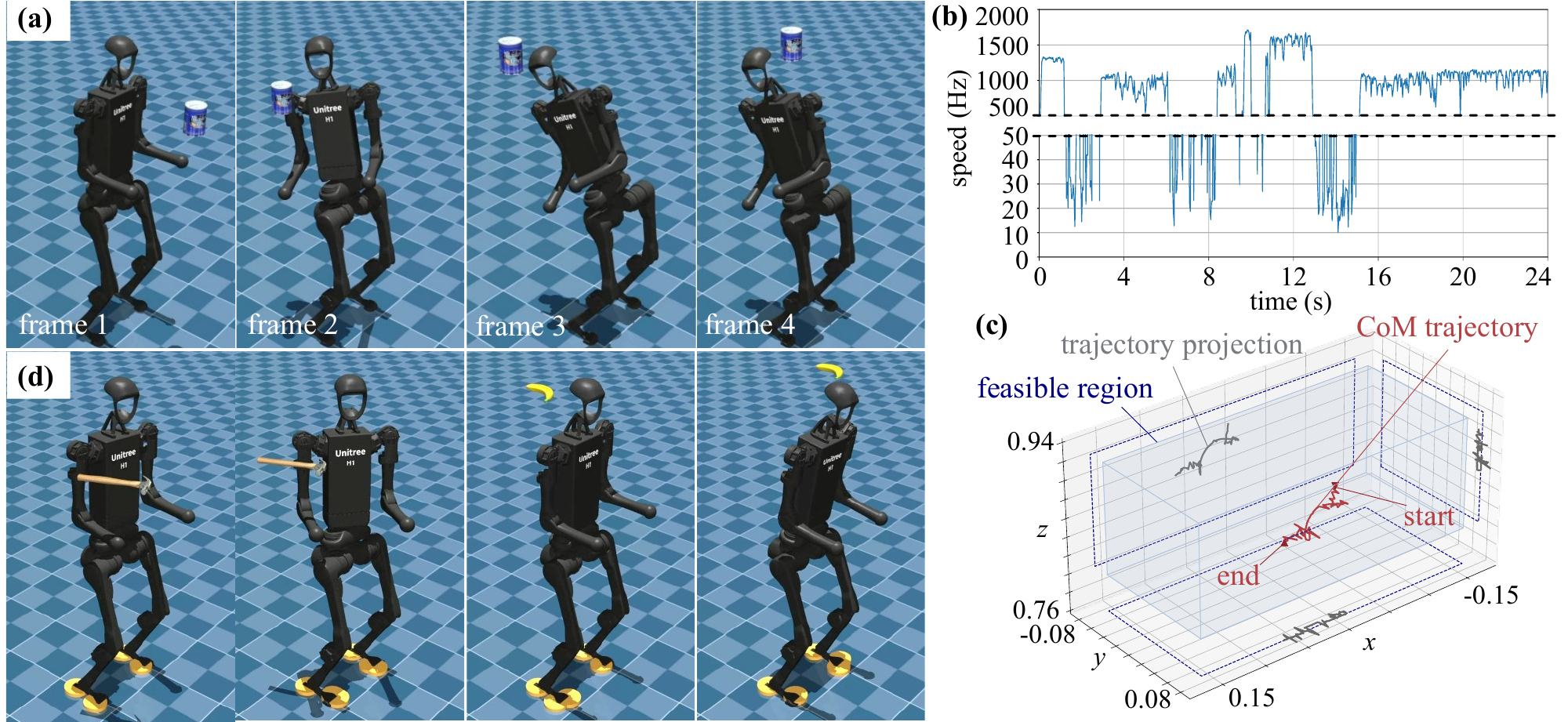}
    \caption{Dynamic obstacle avoidance for humanoid robot. (a) Key frames of the H1 robot avoiding obstacles. (b) Speed of \ac{PDO-IK} in C++ implementation. (c) The trajectory of \ac{CoM} of the robot and its feasible region. (d) Demonstrations of some additional collision avoidance experiments.}
    \label{fig:exp-h1}
\vspace{-5mm}
\end{figure}

We further apply our algorithm on humanoid robot (Fig.~\ref{fig:exp-h1}). In this experiment, we let a can (\texttt{002\_master\_chef\_can} from YCB dataset, its downsampled point cloud contains 168 points) fly to the H1 robot \footnote{Unitree Robotics: \url{https://www.unitree.com/h1/}} in a pre-defined trajectory. The H1 robot is a humanoid robot that contains 19 \ac{DOF}s (5 on each leg, 4 on each arm, 1 on the torso). The H1 robot needs to avoid the can in real time and remains its feet in a fixed position. Moreover, the \ac{CoM} needs to maintain within a feasible region to ensure the stability of the robot. In this experiment, the left ankle of the robot is considered as the base link, which is fixed at $[0.2, 0, 0]^\top$ and the right ankle is considered as the end effector. We fix the right ankle by setting its objective position at $\mathbf{u}^* = [-0.2, 0, 0]^\top$.

For the robot stability, we add constraints to robot \ac{CoM} position $\mathbf{c} \in \mathbb{R}^3$:
\vspace{-2mm}
\begin{equation}
    [-0.16, -0.07, 0.8]^\top < \mathbf{c} < [0.16, 0.075, 0.94]^\top
    \label{eq:com}
\vspace{-2mm}
\end{equation}
where $\mathbf{c}$ is the weighted combination of every links' center position:
\vspace{-1mm}
\begin{subequations}
\vspace{-2mm}
\begin{equation}
    \mathbf{c} = \sum_{i=1}^M m_i \mathbf{c}_i = \sum_{i=1}^M m_i [\mathbf{T}_{{\rm mass}, i}(1,4), \ \mathbf{T}_{{\rm mass}, i}(2,4), \ \mathbf{T}_{{\rm mass}, i}(3,4)]^\top
\vspace{-2mm}
\end{equation}
\vspace{-3mm}
\begin{equation}
    \mathbf{T}_{{\rm mass}, i} = \mathbf{T}_i \cdot ^{i}\mathbf{T}_{{\rm mass}, i}
\vspace{-1mm}
\end{equation}
\end{subequations}
where $m_i$ is the weight of link $i$ and $^{i}\mathbf{T}_{{\rm mass}, i}$ is the transformation matrix of the \ac{CoM} of link $i$ with respect to its own reference frame. After \ac{PDO-IK} solves for a feasible solution $\boldsymbol{\theta}^*$, we use a PD controller to control the motors to reach $\boldsymbol{\theta}^*$.

The algorithm is implemented in C++ on a laptop computer with AMD Ryzen 7 CPU with 16GB RAM. Fig.~\ref{fig:exp-h1}a shows a series of key frames of the simulation experiment on Mujoco \cite{todorov2012mujoco}. The speed of the algorithm varies from 10 to 1500 Hz (Fig.~\ref{fig:exp-h1}b), indicating that the algorithm is capable for real-time collision avoidance in dynamic environments.

The trajectory of the \ac{CoM} is shown in Fig.~\ref{fig:exp-h1}c. The blue box is the feasible region defined by Eq.~\ref{eq:com}. The \ac{CoM} occasionally violates the \ac{CoM} constraints of no more than 0.01m but will quickly return to the feasible region thereafter. Fig.~\ref{fig:exp-h1}d shows some additional demonstrations of the humanoid avoiding dynamic obstacles (\texttt{048\_hammer} and \texttt{011\_banana} from the YCB dataset).

\section{Conclusions and Future Work}\label{sec:conclusion}
We present PDO-IK, a distance-based algorithm for constrained IK problems. It addresses inefficiencies in previous distance-based methods by leveraging the kinematic chain and new formulations for joint limit constraints, collision avoidance constraints, and end effector objectives. Experiments show that our method runs faster, can handle various constraints, and provide more accurate solutions than recent distance-based methods. Finally, experiments on the H1 humanoid robot demonstrate the generalization ability of our method and its usage for collision avoidance in dynamic environments.

Our method has several limitations to be addressed in our future work. First, the angle decomposition process introduces extra DoFs and reduces the speed of the algorithm. Additionally, the collision avoidance constraints assume joints are spherical and links are spheroidal, which might poorly approximate complex robot shapes. Better approximations could be achieved by attaching more points to the robot at the cost of potentially higher computational burden. Moreover, this paper doesn't consider robot self-collision but we believe such constraints can be achieved by constraining the distances between points attached on the robot. Finally, tests on different types of robot structures can be conducted to further broaden the application of the algorithm.

\clearpage
\bibliographystyle{splncs04}
\bibliography{wafr}

\end{document}